# Strategies of Effective Digitization of Commentaries and Sub-commentaries: Towards the Construction of Textual History

**Diptesh Kanojia[1,2,3], Malhar Kulkarni[2], Sayali Ghodekar[2], Eivind Kahrs[4] and Pushpak Bhattacharyya[2]**

[1]IITB-Monash Research Academy, [2]IIT Bombay, [3]Monash University, [4]University of Cambridge
{diptesh, malhar, pb}@iitb.ac.in, sayalighodekar@gmail.com, egk1000@cam.ac.uk

## Abstract

This paper describes additional aspects of a digital tool called the 'Textual History Tool'. We describe its various salient features with special reference to those of its features that may help the philologist digitize commentaries and sub-commentaries on a text. This tool captures the historical evolution of a text through various temporal stages, and interrelated data culled from various types of related texts. We use the text of the *Kāśikāvṛtti* (KV) as a sample text, and with the help of philologists, we digitize the commentaries available to us. We digitize the *Nyāsa* (Ny), the *Padamañjarī* (Pm) and sub-commentaries on the KV text known as the *Tantrapradīpa* (Tp), and the *Makaranda* (Mk). We divide each commentary and sub-commentary into functional units and describe the methodology and motivation behind the functional unit division. Our functional unit division helps generate more accurate phylogenetic trees for the text, based on distance methods using the data entered in the tool.

## 1 Introduction

Texts are invaluable resources of any culture. They provide information that helps a reader understand the intellectual tradition of that culture. In the Indian context, historically speaking, Sanskrit texts were composed and transmitted, orally as well as in written form, for more than at least two thousand years. These texts serve the purpose of sources of Indian culture. Since, these Sanskrit texts usually have long time span of transmission, it affects the very nature of the composition of such texts and their transmission. Many texts are composed in terse *sutra* style, whereas those in elaborate *verse* or *prose* may be prone to loss of meaning comprehension on the part of the contemporary reader. There is another aspect worth mentioning here, since these texts travel in the course of time, they also stand testimony to the evolution and development of the intellectual process.

The fuller comprehension of the terse texts depends on the elaborate explanation of words used in the *sutra*. It is necessary to contain the loss of meaning comprehension of texts which travel in the course of history by codifying a methodology to interpret those texts. When texts travel, they also kickstart a process of thinking based on those texts which causes the development of thought.

All three features mentioned above are the characteristics of what is known as "Commentatorial Texts", popularly known as "Commentaries". Commentaries elaborately explain the wordings of terse *sutras*. They codify the various methodologies of interpreting these texts and they also act as a mirror as far as the process of intellectual development is concerned. When we talk of a specific case like the *Paninian* tradition, we can see that the commentaries such as the *Mahābhāṣya* and the *Kāśikāvṛtti* explain the *Paninian sutras* in detail. They also put in place various methodologies like *anuvṛtti, adhikāra, ekavākyatā,* in order to interpret the *sutras* of *Panini*. They also represent evolution and development of linguistic thought in India over a period of more than one thousand years. It is therefore important to have with us a critical text of the commentary, for example the *Kāśikāvṛtti* (KV), to accurately perform the above-mentioned functions. Interestingly, in the Indian context, all the above-mentioned features of a commentary (KV) with respect to an original *sutra* text, the *Aṣṭādhyāyī* (AST) are also applicable to that same commentary (KV) and its sub-commentaries (Ny and Pm). It is also worth noting that what is applicable to a commentary (KV) and its sub-commentaries (Ny and Pm), is also applicable to the sub-commentaries (Ny and Pm) and their sub-sub-commentaries (Tp and Mk). We also believe that this recurrence continues even today and will continue in the future course of time and, most importantly, all of this will contribute to the understanding and reconstruction of the source text.

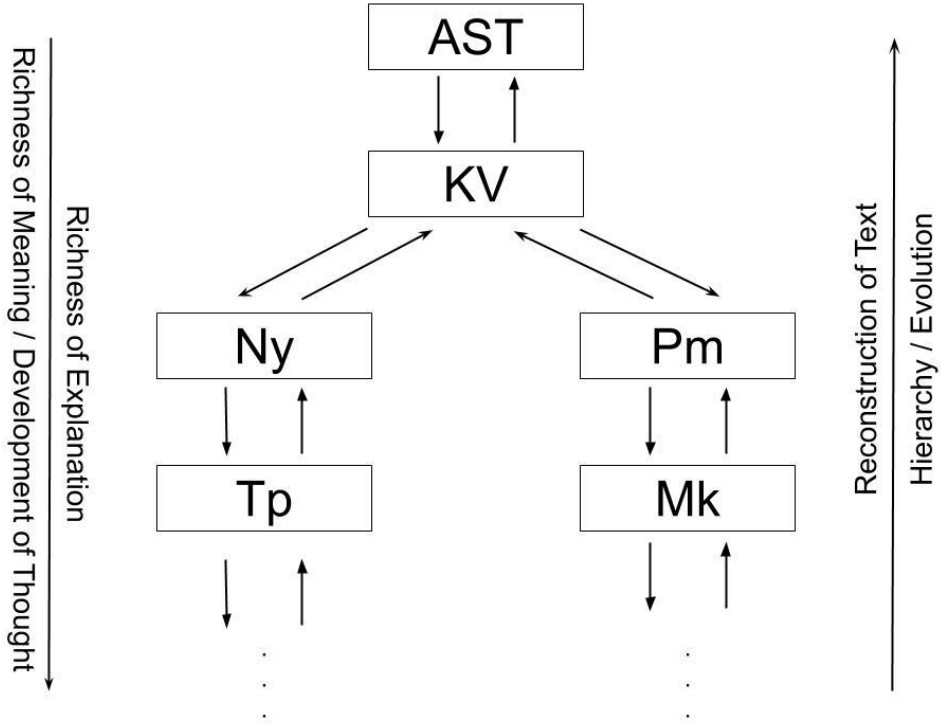

**Figure 1:** The hierarchy of evolution vis-a-vis the development of thought.

The above diagram describes the evolution of commentaries for Panini's *Aṣṭādhyāyī* (A) and depicts the evolution of sub-commentaries such as the Ny and the Pm which are prominent in terms of the evolution of thought, and in terms of richness of explanation. This is observed in the commentaries and sub-commentaries on it. In our tool, we propose a hierarchy of data entry for the manuscript text and its related texts. We require users to enter the manuscript text. We, then, allow addition of commentaries on the text for functional units or *sutras* that have already been entered into the database. Once the commentaries are in the database, we allow the user to choose the commentary for which a sub-commentary is to be added and allow the user the option to enter the sub-commentary text. This recursive approach allows one to maintain the hierarchy and helps preserve the thought development which can be seen while entering data in a sequential manner.

**The key contribution of this paper is**:
- To describe a tool which embodies the strategies to effectively digitize the commentaries and sub-commentaries towards achieving the goal of text reconstruction and its historical evolution.

## 2 Related Work

In this section, we describe the related work divided into two sub-sections, *i.e.*, Computational Related Work, and Philological Related Work.

*2.1 Computational Related Work*

The TEI Critical Edition Toolbox is a tool for preparing a digital TEI critical edition which allows you to check for the encoding of the text. It also facilitates the parallel look-up of the manuscript version by visualizing them on a web-based GUI, although the software is not available for download and offline use yet. In the current state, it accepts only TEI format XML files, but does not allow one to generate versions. A technique for textual criticism is also provided by West (1973). Classical Text Editor allows one to build a critical edition and critical apparatus manually. It also allows one to prepare the genealogical data to be fed to a phylogenetic tree creation program. It allows one to collate the textual versions and edit them in an offline interface.

Dr Sathaye demonstrated work on creating a critical edition of the *Vetālapañcaviṃśati* at the World Sanskrit Conference (WSC) 2018, and at a workshop at the Bhandarkar Oriental Research Institute in 2019. Maas (2010) discusses the creation of a critical edition of *Pātañjalayogaśāstra*. Philips (2010) also discusses the transmission of the *Mahābhārata*, but none of the above discuss a computational tool which can facilitate the entry of sub-commentaries via a digital method. Our work provides this significant edge where a complete thought development and richness of the meaning explained via various scholarly works can be captured in a single interface. Kanojia *et al.* (2019) provide a panoramic view of the Textual History Tool which we enhance with regard to commentaries and sub-commentaries, but they do not include a detailed discussion about sub-commentaries like the one we present in this paper.

*2.2 Philological Related Work*

When we study the texts of the Indian grammatical tradition, that too, the *Paninian* one, traditional commentators like Madhava and Bhattoji Dikshita, etc. (Kulkarni, 2002; Kulkarni and Kahrs, 2015), and modern scholars such as Kielhorn (1887) and Kulkarni (2012) observe that the text of the *Aṣṭādhyāyī* (A) has evolved in the course of time. The text of the *sutras* that Patanjali had in front of him is not the same as we have it today. As shown by Kulkarni (2016), the traditional commentators quoted above, consider the text of the KV as an important stage of evolution of the text of the A because the KV brought about numerous modifications in the text of the A, by sometimes adding a word or two in the *sutra*, splitting one *sutra* into two, converting a later *vārttika* into a *sutra*, *etc.* Joshi *et al.* (1995) state that the KV also preserved a tradition of interpretation of the AST, independent of *Patanjali*. Bronkhorst (2009) showed that the KV also has an interface with other, non-paninian, Sanskrit grammatical traditions.[1]

In order to study this stage of evolution further, when we turn to the printed text of the KV as available to us through more than ten editions, as of now, we notice that the printed editions do not present to us a picture of a uniform text, and rather suggest that this text of the KV that we have with us today must have evolved in a particular manner, historically. Kulkarni (2012) studied the '*gaṇapāṭhas*' and after analyzing the data from manuscripts showed how the number of words in a '*gaṇa*' increased in the course of time and formulated the stages of this historical development.[2]

---

[1] There is no evidence that the KV was ever handed down orally. So oral transmission cannot be used as a resource in the reconstruction of the evolution of the KV. A modern counterexample will also make this point clearer: The text of the *Vaiyākaraṇasiddhāntakaumudī* (SK) was handed down orally, and even Malhar Kulkarni memorised it as part of his traditional education. In fact, it can be said that the primary focus of the structure of the text of the SK is for oral transmission.

## 3 Textual History Tool

We demonstrated an initial version of the Textual History Tool at the WSC 2018 where the tool had capabilities of taking various forms of input from annotators and storing the manuscript data. An initial description of the tool with its technical architecture and schematic representation is described by Kanojia *et al.* (2018, 2020). Another paper describing the Textual History Tool (Kanojia *et al.* 2019) was published at the ISCLS 2019 with many modifications based on a lot of beta testing and feature addition. In this regard, the tool was demonstrated again at the ISCLS 2019 with features such as phylogenetic tree generation based on the manuscript versions. This enhanced version of the tool also contained features for adding commentaries on the text. Since then, the tool has added functionalities such as sub-commentary addition to the commentary data and allows annotators to edit those already added in the database. We provide further details and changes made in the tool since then. We provide the step by step methodology of adding/editing commentaries below:

- Click on the "Add" button to add a new commentary based on the existing *sutra*. A user can add up to eight commentaries as the number of commentaries are dynamic and to be decided by the user depending on the number of functional units.
- Display the commentary text along with the label and type of evidence if it exists in the database.
- Add the Commentary label (eg: Ny, Pm) and commentary text.
- Option to mark the evidence in the commentary as Direct, Indirect, both or default. The direct and indirect evidences are further divided into various subtypes.
- Option to add sub-commentaries using a button, depending on the functional units of the text.

Our current contribution is mainly adding the sub-commentaries to the commentary data and providing options for adding and editing them, as per the user's expertise. We use this additional feature to add the sub-commentary data for the KV text and substantiate our claim that the tool can indeed be used to add and edit sub-commentaries. Screenshots of the said function are provided below in Figure 2 and 3.

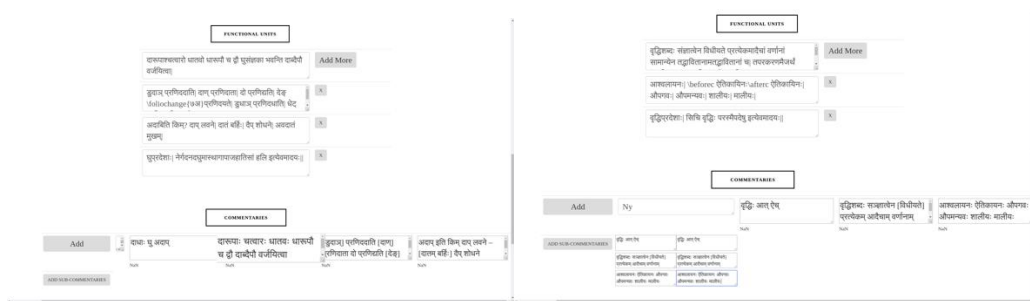

**Figures 2 and 3**: Display tool functionality to add commentaries and tool functionality to add sub-commentaries, respectively.

### 3.1 Classification of the Textual Evidence from Commentaries

Textual evidence as available from commentaries can be classified into Direct and Indirect evidence as found through the historical mentions and explanations of the text being discussed. Indirect evidence can further be classified into various different categories as shown in figure 4 below. This classification was proposed by Kulkarni and Kahrs (2019).

---
[2] Also, Kulkarni presented another paper at the WSC 2018 studying in detail the printed editions of the KV on various '*gaṇas'* (accepted for publication).

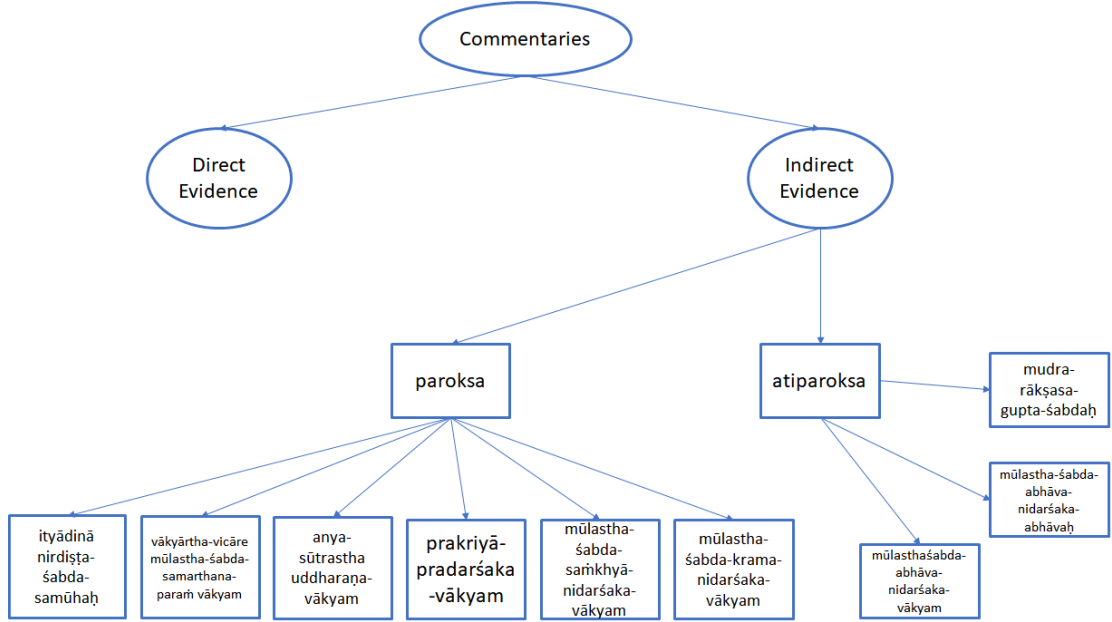

**Figure 4**: The classification of evidence found in commentaries as proposed by Kulkarni and Kahrs (2019)

## 4 Reconstruction of the History of the Text

We provide below two detailed examples of how commentaries and sub-commentaries may explain the text better. These example *sutras* are taken from printed editions of the Ny and Pm, and the Ny and Tp.

### 4.1 Example 1

| 1.1.1. | *Sutra* | वृद्धिः आत् ऐच् |
|---|---|---|
| 1.1.1.1. | Introduction and Meaning | सञ्ज्ञात्वेन [विधीयते] प्रत्येकम् आदैचाम् वर्णानाम् सामान्येन तद्भावितानाम् [अतद्भावितानाम्] च तपरकरणम् [ऐजर्थम्] तात् अपि परः तपरः इति खट्वैडकादिषु [त्रिमात्रचतुर्मात्रप्रसङ्गनिवृत्तये] |
| 1.1.1.2. | Examples | आश्वलायनः ऐतिकायनः औपगवः औपमन्यवः शालीयः मालीयः |
| 1.1.1.3. | Other Occurrences of the term | वृद्धिप्रदेशाः सिचि वृद्धिः परस्मैपदेषु इति एवमादयः |

**Table 1:** Example of a *sutra* from KV (*Sutra* 1.1.1.)

It is observed that amongst the three sections of the KV 1.1.1.3. has got no support in the form of evidence from the Ny as well as the Pm. In the remaining two sections, there are twenty-five words out of which twenty-four are supported by the evidence found in the Ny and twelve are supported by the evidence found in the Pm. We note that the evidence found in the Ny and the one found in the Pm is uniform and has no variation to offer. Based on this, we can say that the transmission of the two sections of the KV on AST 1.1.1. can be reconstructed in near entirety on the basis of the evidence available to us from the Ny supported by the Pm. A comparison of this uniform evidence with the manuscript evidence may give rise to the knowledge of uniform transmission of the two sections of the KV on A 1.1.1. However, the same comparison will also reveal the fact that 1.1.1.3. which has no support whatsoever from either the Ny or the Pm, receives concrete support from the manuscripts. This leads to the variation in the transmission of the text of 1.1.1.3. This might suggest the fact that the archetype of the manuscripts that we consult on 1.1.1.3. could be post-Pm. This aspect is treated in detail by Kulkarni and Kahrs (forthcoming).

**4.2 Example 2**

| 2.1.22. | *Sutra* | तत्पुरुषः |
|---|---|---|
| 2.1.22.1. | Introduction and Meaning | तत्पुरुषः इति संज्ञा ऽधिक्रियते प्राग् बहुव्रीहेः। यानित ऊर्ध्वम् अनुक्रमिष्यामः, तत्पुरुषसंज्ञास्ते वेदितव्याः। |
| 2.1.22.2. | Examples | वक्ष्यति, द्वितीय श्रितातीतपतित इति। कष्टश्रितः। पूर्वाचार्यसंज्ञा चेयं महती, तदङ्गीकरणौपाधेरपि तदीयस्य परिग्रहार्थम्, उत्तरपदार्थप्रधानस् तत्पुरुषः इति। |
| 2.1.22.3. | Other Occurrences of the term | तत्पुरुषप्रदेशाः तत्पुरुषे कृति बहुलम् इत्येवम् आदयः। |

**Table 2:** Example of a sutra from the KV (Sutra 2.1.22.)

In Example 2, we present the text of the KV on AST 2.1.22. This is also a *saṁjñā sutra* like A 1.1.1 so the text of the KV is also divided into three sections as above. We observe that the Ny presents evidence that supports the existence of nine words in the first two sections. We also note that the Tp presents evidence to support only one word out of these nine, but without any variation. So, the textual transmission of these two sections can be said to be uniform, extending itself to even the Tp. Even here the Ny does not offer any evidence to support the existence of 2.1.22.3, and we also note that the Tp too does not offer any evidence to support the existence of 2.1.22.3. So, we can say that overall, in all the sections of 2.1.22, a uniform transmission is observed as far as the Ny and the Tp are concerned. As far as 2.1.22.3 is concerned, in comparison with comments made above, we can extend the argument and say that our archetype can be said to be post-Tp. So, a collection of evidence from the Tp on similar sections in all the available *saṁjñā sutras* can help shed more light on the transmission of this section.

## 5 Conclusion and Future Work

Our work describes an effective method for the digitization of commentaries and sub-commentaries available for a text. We describe a digital tool which provides the access to users based on an authentication-based platform and helps them add sub-commentary data in a hierarchical and recursive manner. We highlight the changes made to this Textual History Tool presented earlier by Kanojia *et al.* (2019), and show, with the help of screenshots, how improvements made to the tool functionalities help annotators digitize the commentary data. Using this approach one can effectively digitize the available data and use them to create a phylogenetic tree for the text. This is a novel contribution of our work done with the Textual History Tool. The tree can be created by using the various versions of the text as available in the manuscripts and including commentaries or by only using the commentary data. The tree generation methodologies which have been described by Kanojia *et al.* (2019) are the same for our work. We also present examples of sub-commentary data described in the paper. The trees which are generated via our approach assume that the commentaries present in the form of printed editions are authorized versions of the written text. We aim to use this tool and the work done to successfully help digitize many texts and commentaries over them.

In future, we would like to include the testimonia addition to the tool. This can help create trees even based on mentions of the text in other texts. These could belong to the same tradition or some other tradition. This can help provide clues to the tree generation algorithms as to when and where the text was mentioned and what form of the text was quoted in the other text. We also aim to extend our work to other texts than the KV and show that the tool has indeed a generic applicability towards text digitization and reconstruction.

## Acknowledgement

We acknowledge the efforts of our annotators Dr Nilesh Joshi and Dr Irawati Kulkarni who have helped us digitize the text of the KV.

## Bibliographical References